\documentclass[letterpaper]{article} 
\usepackage{aaai23}  
\usepackage{times}  
\usepackage{helvet}  
\usepackage{courier}  
\usepackage[hyphens]{url}  
\usepackage{graphicx} 
\usepackage{natbib}  
\usepackage{caption} 
\usepackage{algorithm}
\usepackage{multirow}
\usepackage{multicol}
\usepackage{amsmath}

\usepackage{algorithmicx,algorithm}
\usepackage[noend]{algpseudocode}
\usepackage{amsmath}
\usepackage{color}
\usepackage{diagbox}
\usepackage{amssymb}
\usepackage[table]{xcolor}
\usepackage{color}

\newcommand{\minorimprove}[1]{\textcolor{blue}{#1}}
\newcommand{\largeimprove}[1]{{\textcolor{red}{#1}}}
\usepackage{marvosym}

\usepackage{newfloat}
\usepackage{listings}
\DeclareCaptionStyle{ruled}{labelfont=normalfont,labelsep=colon,strut=off} 
\floatstyle{ruled}
\newfloat{listing}{tb}{lst}{}
\floatname{listing}{Listing}
\title{Curriculum Temperature for Knowledge Distillation}
\author{
    Zheng Li \textsuperscript{\rm 1},
    Xiang Li \textsuperscript{\rm 1}\thanks{Corresponding Author. This work is partially done when Zheng Li is an intern at Megvii.},
    Lingfeng Yang \textsuperscript{\rm 2},
    Borui Zhao \textsuperscript{\rm 3}, \\
    Renjie Song \textsuperscript{\rm 3},
    Lei Luo \textsuperscript{\rm 2},
    Jun Li \textsuperscript{\rm 2},
    Jian Yang \textsuperscript{\rm 1}\footnotemark[1]
}
\affiliations{
    \textsuperscript{\rm 1} Nankai University
    \textsuperscript{\rm 2} Nanjing University of Science and Technology
    \textsuperscript{\rm 3} Megvii Technology \\
    zhengli97@mail.nankai.edu.cn, \{xiang.li.implus, csjyang\}@nankai.edu.cn,\\ zhaoborui.gm@gmail.com, songrenjie@megvii.com, \{yanglfnjust, cslluo, junli\}@njust.edu.cn
}

\begin{document}

\maketitle
\begin{abstract}

Most existing distillation methods ignore the flexible role of the temperature in the loss function and fix it as a hyper-parameter that can be decided by an inefficient grid search. In general, the temperature controls the discrepancy between two distributions and can faithfully determine the difficulty level of the distillation task. Keeping a constant temperature, i.e., \emph{a fixed level of task difficulty}, is usually sub-optimal for a growing student during its progressive learning stages. In this paper, we propose a simple curriculum-based technique, termed \textbf{C}urriculum \textbf{T}emperature for \textbf{K}nowledge \textbf{D}istillation~(\textbf{CTKD}), which controls the task difficulty level during the student's learning career through a dynamic and learnable temperature.
Specifically, following an easy-to-hard curriculum, we gradually increase the distillation loss w.r.t. the temperature, leading to increased distillation difficulty in an adversarial manner.
As an easy-to-use plug-in technique, CTKD can be seamlessly integrated into existing knowledge distillation frameworks and brings general improvements at a negligible additional computation cost. Extensive experiments on CIFAR-100, ImageNet-2012, and MS-COCO demonstrate the effectiveness of our method. Our code is available at~\url{https://github.com/zhengli97/CTKD}.

\end{abstract}

\section{Introduction}

Knowledge distillation~\cite{hinton2015distilling}~(KD) has received increasing attention from both academic and industrial researchers in recent years.
It aims at learning a comparable and lightweight student by transferring the knowledge from a pretrained heavy teacher. The traditional process is implemented by minimizing the KL-divergence loss between two predictions obtained from the teacher/student model with a fixed temperature in the softmax layer. As depicted in~\cite{hinton2015distilling,liu2022meta,chandrasegaran2022revisiting}, the temperature controls the smoothness of distribution and can faithfully determine the difficulty level of the loss minimization process. 
Most existing works~\cite{tung2019similarity,chen2020online,ji2021refine} ignore the flexible role of the temperature and empirically set it to a fixed value (e.g., 4). Differently, MKD~\cite{liu2022meta} proposes to learn the suitable temperature via meta-learning. However, it has certain limitations that require an additional validation set to train the temperature module, which complicates the training process. Besides, it mainly focuses on the strong data augmentation condition, neglecting that most existing KD methods work under normal augmentation. Directly combining MKD with existing distillation methods under strong augmentation may cause severe performance degradation~\cite{das2020empirical}.

In human education, teachers always train students with simple curricula, which start from easier knowledge and gradually present more abstract and complex concepts when students grow up. This curriculum learning paradigm has inspired various machine learning algorithms~\cite{caubriere2019curriculum,duan2020curriculum}. In knowledge distillation, LFME~\cite{xiang2020learning} adopt the classic curriculum strategy and propose to train the student gradually using samples ordered in an easy-to-hard sequence. RCO~\cite{jin2019knowledge} propose to utilize the sequence of the teacher's intermediate states as the curriculum to gradually guide the learning of a smaller student. 
The progressive curricula based on data samples and models can help students learn better representations during distillation, but it requires a careful curriculum design and complex computational process, making it hard to deploy into existing methods.

In this paper, we propose a simple and elegant curriculum-based approach, called Curriculum Temperature for Knowledge Distillation~(CTKD), which enhances the distillation performance by progressively increasing the learning difficulty level of the student through a dynamic and learnable temperature. 
The temperature is learned during the student's training process with a \emph{reversed} gradient that aims to maximize the distillation loss~(i.e., increasing the learning difficulty) between teacher and student in an adversarial manner. Specifically, the student is trained under a designed curriculum via the learnable temperature: following the easy-to-hard principle, we gradually increase the distillation loss w.r.t. the temperature, resulting in increased learning difficulty through simply adjusting the temperature dynamically. This operation can be easily implemented by a non-parametric gradient reversal layer~\cite{ganin2015unsupervised} to reverse the gradients of the temperature, which hardly introduces extra computation budgets. Furthermore, based on the curriculum principle, we explore two~(global and instance-wise) versions of the learnable temperature, namely Global-T and Instance-T respectively. As an easy-to-use plug-in technique, CTKD can be seamlessly integrated into most existing state-of-the-art KD frameworks and achieves comprehensive improvement at a negligible additional computation cost. %

In summary, our contributions are as follows: 
\begin{itemize}
    \item We propose to adversarially learn a dynamic temperature hyperparameter during the student's training process with a reversed gradient that aims to maximize the distillation loss between teacher and student. 
    \item We introduce simple and effective curricula which organize the distillation task from easy to hard through a dynamic and learnable temperature parameter. 
    \item Extensive experiment results demonstrate that CTKD is a simple yet effective plug-in technique, which consistently improves existing state-of-the-art distillation approaches with a substantial margin on CIFAR-100 and ImageNet.
\end{itemize}

\section{Related Work}

\textbf{Curriculum Learning.} Originally proposed by~\cite{bengio2009curriculum}, curriculum learning~\cite{wang2021survey} is a way to train networks by organizing the order in which tasks are learned and incrementally increasing the learning difficulty~\cite{morerio2017curriculum,caubriere2019curriculum}. 
This training strategy has been widely applied in various domains, such as computer vision~\cite{wu2018learning,sinha2020curriculum} and natural language processing~\cite{platanios2019competence,tay2019simple}.
Curriculum Dropout~\cite{morerio2017curriculum} dynamically increases the dropout ratios in order to improve the generalization ability of the model. PG-GANs~\cite{karras2017progressive} learn to sequentially generate images from low-resolution to high-resolution, and also grew both generator and discriminator simultaneously.
In knowledge distillation, various works~\cite{xiang2020learning,zhao2021knowledge} adopt the curriculum learning strategy to train the student model. LFME~\cite{xiang2020learning} proposes to use the teacher as a difficulty measure and organize the training samples from easy to hard so that the model can receive a less challenging schedule. RCO~\cite{jin2019knowledge} proposes to utilize the sequence of teachers' intermediate states as a curriculum to supervise the student at different learning stages.

\noindent\textbf{Knowledge Distillation.} KD~\cite{hinton2015distilling} aims at effectively transferring the knowledge from a pretrained teacher model to a compact and comparable student model. 
Traditional methods propose to match the output distributions of two models by minimizing the Kullback-Leibler divergence loss with a fixed temperature hyperparameter. To improve distillation performance, existing methods have designed various forms of knowledge transfer. It can be roughly divided into three types, logit-based~\cite{chen2020online,li2020online,zhao2022decoupled}, representation-based~\cite{yim2017gift,chen2021distilling} and relationship-based~\cite{park2019relational,peng2019correlation} methods.
The temperature controls the smoothness of probability distributions and can faithfully determine the difficulty level of the distillation process. As discussed in~\cite{chandrasegaran2022revisiting,liu2022meta}, a lower temperature will make the distillation pays more attention to the maximal logits of teacher output. On the contrary, a higher value will flatten the distribution, making the distillation focus on the logits.
Most works ignore the effectiveness of the temperature on distillation and fix it as a hyperparameter that can be decided by an inefficient grid search. However, keeping a constant value, i.e., a fixed level of distillation difficulty, is sub-optimal for a growing student during its progressive learning stages.

Recently, MKD~\cite{liu2022meta} proposes to learn the temperature by performing meta-learning on the extra validation set. It mainly works on the ViT~\cite{dosovitskiy2020image} backbone with strong data augmentation while most existing KD methods work under normal augmentation. Directly applying MKD to other distillation methods may weaken the effect of distillation~\cite{das2020empirical}.
Our proposed CTKD is more efficient than MKD since we don't need to pay the effort to split and preserve an extra validation set. Besides, CTKD works under normal augmentation, so it can be seamlessly integrated into existing KD frameworks. The detailed comparison and discussion are attached in the supplement.

\section{Method}

In this section, we first review the concept of knowledge distillation and then introduce our proposed curriculum temperature knowledge distillation technique.

\begin{figure*}[t]
	\centering 
	\includegraphics[width=1\linewidth]{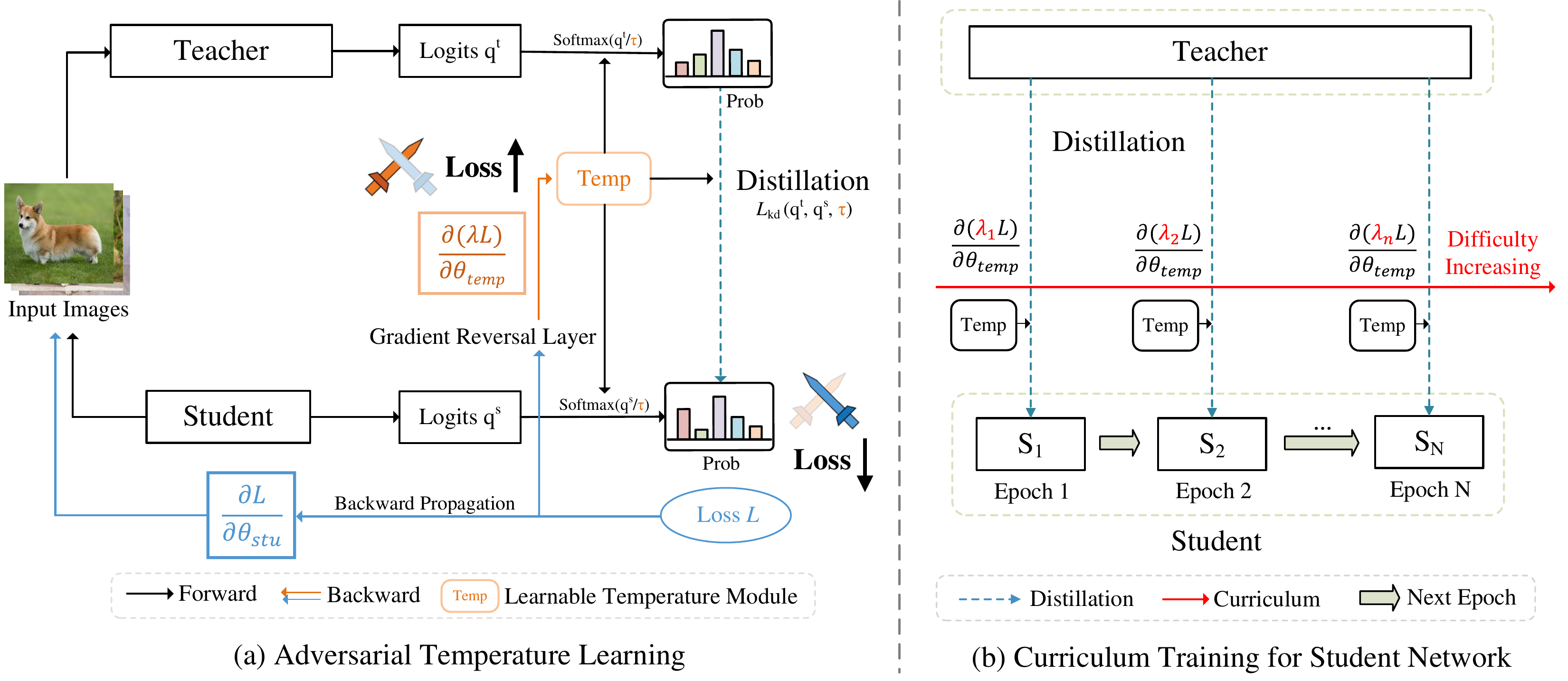}
	\caption{
		An overview of our proposed Curriculum Temperature for Knowledge Distillation~(CTKD). (a) We introduce a learnable temperature module that predicts a suitable temperature $\tau$ for distillation. The gradient reversal layer is proposed to reverse the gradient of the temperature module during the backpropagation.
		(b) Following the easy-to-hard curriculum, we gradually increase the parameter $\lambda$, leading to increased learning difficulty w.r.t. temperature for the student.
	}
	\label{fig:arch_compare}
\end{figure*}

\subsection{Background}

Knowledge distillation~\cite{hinton2015distilling}, as one of the main network compression techniques, has been widely used in many vision tasks~\cite{liu2019structured,ye2019student,li2021online,li2022knowledge}. The traditional two-stage distillation process usually starts with a pre-trained cumbersome teacher network. Then a compact student network will be trained under the supervision of the teacher network in the form of soft predictions or intermediate representations~\cite{romero2014fitnets,yim2017gift}. After the distillation, the student can master the expertise of the teacher and use it for final deployment.
Given the labeled classification dataset $D=\{({x}_{i},{y}_{i})\}_{i=1}^{I}$, the Kullback-Leibler (KL) divergence loss is used to minimize the discrepancy between the soft output probabilities of the student and teacher model:
\begin{equation}
L_{kd}(q^{t}, q^{s}, \tau) = \sum_{i=1}^{I}\tau^{2} KL(\sigma (q^{t}_{i}/\tau),\sigma (q^{s}_{i}/\tau)),
\label{equation:kd}
\end{equation}
where $q^{t}$ and $q^{s}$ denote the logits produced by teacher and student, $\sigma(\cdot)$ is the softmax function, and $\tau$ is the temperature to scale the smoothness of two distributions. 
As discussed in previous works~\cite{hinton2015distilling,liu2022meta}, a lower $\tau$ will sharpen the distribution, enlarge the difference between two distributions and make distillation focus on the maximal logits of teacher prediction. While a higher $\tau$ will flatten the distribution, narrow the gap between two models and make the distillation focus on whole logits.
Therefore, the temperature value $\tau$ can faithfully determine the difficulty level of the KD loss minimization process by affecting the probability distribution.

\subsection{Adversarial Distillation}

For a vanilla distillation task, the student $\theta_{stu}$ is optimized to minimize the task-specific loss and distillation loss. The objective of the distillation process can be formulated as follows:

\begin{figure*}[t]
	\centering 
	\includegraphics[width=0.7\linewidth]{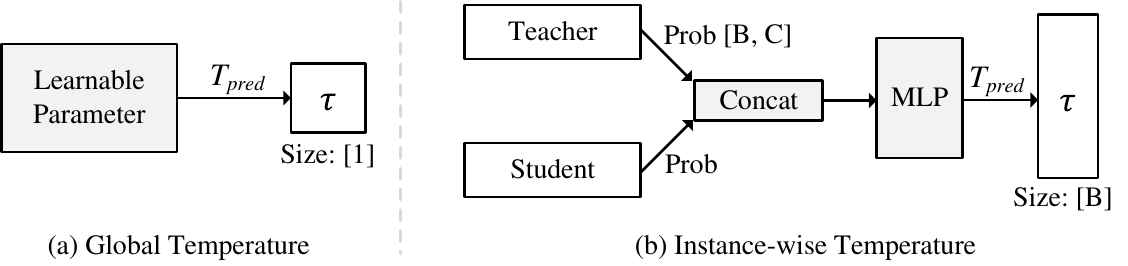}
	\caption{
		The illustrations of global and instance-wise temperature modules. $B$ denotes the batch size, $C$ denotes the number of classes. $\tau$ is the final temperature.
	}
	\label{fig:temp}
\end{figure*}

\begin{equation}
\begin{aligned}
    \underset{\theta_{stu}}{\min}~ L(\theta_{stu}) & = 
    \underset{\theta_{stu}}{\min}~ \sum_{x \in D} \alpha_{1}L_{task}\left(f^{s}(x; \theta_{stu}), y \right) 
    \\ & +\alpha_{2}L_{kd}\left(f^{t}(x; \theta_{tea}), f^{s}(x; \theta_{stu}), \tau \right).
\label{equation:vanilla_loss}
\end{aligned}
\end{equation}
where $L_{task}$ is the regular cross-entropy loss for the image classification task, $f^{t}(\cdot)$ and $f^{s}(\cdot)$ denotes the function of teacher and student. $\alpha_{1}$ and $\alpha_{2}$ are balancing weights.

In order to control the learning difficulty of the student via dynamic temperature,
inspired by GANs~\cite{goodfellow2014generative}, we propose to adversarially learn a dynamic temperature module $\theta_{temp}$ that predicts a suitable temperature value $\tau$ for the current training. This module is optimized in the opposite direction of the student, intending to maximize the distillation loss between the student and teacher. 
Different from vanilla distillation, the student $\theta_{stu}$ and temperature module $\theta_{temp}$ play the two-player mini-max game with the following value function $L(\theta_{stu}, \theta_{temp})$:
\begin{equation}
\begin{aligned}
    & \underset{\theta_{stu}}{\min}~\underset{\theta_{temp}}{\max}~ L(\theta_{stu}, \theta_{temp})  
    \\ & = \underset{\theta_{stu}}{\min}~\underset{\theta_{temp}}{\max}~ \sum_{x \in D} \alpha_{1}L_{task}\left( f^{s}(x;\theta_{stu}), y \right) 
    \\ & + \alpha_{2}L_{kd}\left(f^{t}(x; \theta_{tea}), f^{s}(x; \theta_{stu}), \theta_{temp} \right).
\label{equation:ctkd_loss}
\end{aligned}
\end{equation}

We apply the alternating algorithm to solve the problem in Eqn.~\eqref{equation:ctkd_loss}, fixing one set of variables and solving for the other set. Formally, we can alternate between solving these two subproblems:
\begin{equation}
    \hat{\theta}_{stu} = \arg~\underset{\theta_{stu}}{\min}~ L(\theta_{stu}, \hat{\theta}_{temp}),
\label{equation:stu_gradient}
\end{equation}
\begin{equation}
    \hat{\theta}_{temp} = \arg~\underset{\theta_{temp}}{\max}~ L(\hat{\theta}_{stu}, \theta_{temp}).
\label{equation:temp_gradient}
\end{equation}

The optimization process for Eqn.~\eqref{equation:stu_gradient} and Eqn.~\eqref{equation:temp_gradient} can be conducted via stochastic gradient descent~(SGD). The student $\theta_{stu}$ and temperature module $\theta_{temp}$ parameters are updated as follows:
\begin{equation}
    \theta_{stu} ~ \leftarrow ~ \theta_{stu} - \mu \frac{\partial L}{\partial \theta_{stu}},
\label{equation:stu_param}
\end{equation}
\begin{equation}
    \theta_{temp} ~ \leftarrow ~ \theta_{temp} + \mu \frac{\partial L}{\partial \theta_{temp}}.
\label{equation:temp_param}
\end{equation}
where $\mu$ is the learning rate.

In practice, we implement the above adversarial process (i.e., Eqn.~\eqref{equation:temp_param}) by a non-parametric Gradient Reversal Layer~(GRL)~\cite{ganin2015unsupervised}. The GRL is inserted between the softmax layer and the learnable temperature module, as shown in Fig.~\ref{fig:arch_compare}(a). 

\subsection{Curriculum Temperature}


Keeping a constant learning difficulty is sub-optimal for a growing student during its progressive learning stages. In school, human teachers always teach students with curricula, which start with basic~(easy) concepts, and then gradually present more advanced~(difficult) concepts when students grow up. Humans will learn much better when the tasks are organized in a meaningful order.

Inspired by curriculum learning~\cite{bengio2009curriculum}, we further introduce a simple and effective curriculum which organizes the distillation task from easy to hard via directly scaling the loss $L$ by magnitude $\lambda$ w.r.t. the temperature, i.e., $L \rightarrow \lambda L$.
Consequently, the $\theta_{temp}$ would be updated by:
\begin{equation}
    \theta_{temp} ~ \leftarrow ~ \theta_{temp} + \mu \frac{\partial (\lambda L)}{\partial \theta_{temp}}.
\label{equation:ctkd_temp_param}
\end{equation}

At the beginning of training, the junior student has limited representation ability and requires to learn basic knowledge.
We set the initial $\lambda$ value to $0$ so that the junior student can focus on the learning task without any constraints. By gradually increasing $\lambda$, the student learns more advanced knowledge as the distillation difficulty increases.
Specifically, following the basic concept of curriculum learning, our proposed curriculum satisfies the following two conditions: 

(1) Given the unique variable $\tau$, the distillation loss w.r.t. the temperature module~(simplified as $L_{kd}(\tau)$) gradually increases, i.e.,
\begin{equation}
    L_{kd}(\tau_{n+1}) \geq L_{kd}(\tau_{n}),
\label{equation:curriculum_growing}
\end{equation}

(2) The value of $\lambda$ increases, i.e.,
\begin{equation}
    \lambda_{n+1} \geq \lambda_{n}.
\end{equation}
where $n$ represents the n-th step of training. 

In our method, when training at $E_{n}$ epoch, we gradually increase $\lambda$ with a cosine schedule as follows:
\begin{equation}
\begin{aligned}
    \lambda_{n} & = \lambda_{min} \\ & + \frac{1}{2}(\lambda_{max}-\lambda_{min})(1+\cos((1+\frac{\min(E_{n}, E_{loops})}{E_{loops}})\pi).
\label{equation:lambda}
\end{aligned}
\end{equation}
where $\lambda_{max}$ and $\lambda_{min}$ are ranges for $\lambda$. $E_{loops}$ is the hyper-parameter that gradually varies the difficulty scale $\lambda$. In our method, we default to set $\lambda_{max}$, $\lambda_{min}$ and $E_{loops}$ to $1$, $0$ and $10$, respectively. This curriculum indicates that the parameter $\lambda$ increases from $0$ to $1$ during $10$ epochs of training and keeps $1$ until the end. Detailed ablation studies are conducted in Table~\ref{table:curriculum_params} and Table~\ref{table:curriculum_strategy}.

\begin{algorithm}[t]
	\caption{Curriculum Temperature Distillation}
	\hspace*{0.02in} {\bf Input:}
	Training dataset $D=\{({x}_{i},{y}_{i})\}_{i=1}^{I}$; Total training Epoch $N$; Pre-trained Teacher $\theta_{tea}$; Learnable Temperature Module $\theta_{temp}\in\{\theta_{Global}, \theta_{Instance}\}$; \\
	\hspace*{0.02in} {\bf Output:}
	Well-trained Student $\theta_{stu}$; \\
	\hspace*{0.02in} {\bf Initialize:}
	Epoch $n$=1; Randomly initialize $\theta_{stu}$, $\theta_{temp}$; 
	
	\begin{algorithmic}[1]
		\While {$n$ $\leq$ $N$}
        \For{data batch $x$ in D}
        \State Forward propagation through $\theta_{tea}$ and $\theta_{stu}$ to obtain predictions $f^{t}(x; \theta_{tea})$, $f^{s}(x; \theta_{stu})$;
        \State Obtain temperature $\tau$ by $\theta_{temp}$ in Eqn.~\eqref{equation:temp} and parameter $\lambda_{n}$ in Eqn.~\eqref{equation:lambda};
        \State Calculate the loss $L$ and update $\theta_{stu}$ and $\theta_{temp}$ by backward propagation as Eqn.~\eqref{equation:stu_param} and Eqn.~\eqref{equation:ctkd_temp_param};
        \EndFor
        \State \textbf{end for}
		\State $n$=$n$+1;
		\EndWhile
		\State \textbf{end while}
	\end{algorithmic}
	\label{algo:demo}
\end{algorithm}


\begin{table*}[t]
	\begin{center}
		\resizebox{1\linewidth}{!}{
		\begin{tabular}{cccccccccccc}
			\hline\noalign{\smallskip}
			Teacher	    & RN-56  & RN-110 & RN-110 & WRN-40-2 & WRN-40-2 & VGG-13 & WRN-40-2 & VGG-13 & RN-50 & RN-32x4 & RN-32x4 \\
			Acc		    & 72.34  & 74.31  & 74.31  & 75.61    & 75.61    & 74.64  & 75.61    & 74.64  & 79.34 & 79.42   & 79.42  \\
			\hline\noalign{\smallskip}
			Student	    & RN-20  & RN-32  & RN-20 & WRN-16-2 & WRN-40-1 & VGG-8 & SN-V1 & MN-V2 & MN-V2 & SN-V1  & SN-V2  \\
			Acc		    & 69.06  & 71.14  & 69.06 & 73.26    & 71.98    & 70.36 & 70.50 & 64.60 & 64.60 & 70.50  & 71.82  \\
			\hline\noalign{\smallskip}
			Vanilla KD          & 70.66  & 73.08  & 70.66 & 74.92  & 73.54  & 72.98  & 74.83  & 67.37 & 67.35 & 74.07  & 74.45  \\
			\multirow{2}*{CTKD} & 71.19  & 73.52  & 70.99 & 75.45  & 73.93  & 73.52  & 75.78  & 68.46 & 68.47 & 74.48  & 75.31  \\
			~                   & (+0.53) & (+0.44) & (+0.33) & (+0.53) & (+0.39) & (+0.54) & (+0.95) & (+1.09) & (+1.12) & (+0.41) & (+0.86) \\
			\hline
		\end{tabular}
		}
		\caption{
			Top-1 accuracy of the student network on CIFAR-100.
		}
		\label{table:vanilla_kd_cifar100}
	\end{center}
\end{table*}

\subsection{Learnable Temperature Module} 

In this section, we introduce two versions of the learnable temperature module, namely Global-T and Instance-T. 

\textbf{Global-T.} The global version consists of only one learnable parameter, predicting one value $T_{pred}$ for all instances,  as shown in Fig.~\ref{fig:temp}(a). This efficient version does not bring additional computational costs to the distillation process since it only involves a single learnable parameter. 

\textbf{Instance-T.} To achieve a better distillation performance, one global temperature is not accurate enough for all instances. We further explore the instance-wise variant, termed Instance-T, which predicts a temperature for all instances \emph{individually}, e.g., for a batch of 128 samples, we predict 128 corresponding temperature values.
Inspired by GFLv2~\cite{li2020generalized,li2021generalized}, we propose to utilize the statistical information of probability distribution to control the smoothness of itself. 
Specifically, a 2-layer MLP is introduced in our work, which takes two predictions as input and outputs predicted value $T_{pred}$, as shown in Fig.~\ref{fig:temp}(b). During training, the module will automatically learn the implicit relationship between original and smoothed distribution.

To ensure the non-negativity of the temperature parameter and keep its value within a proper range, we scale the predicted $T_{pred}$ with the following equation:
\begin{equation}
    {\tau}=\tau_{init}+\tau_{range}(\delta(T_{pred})).
\label{equation:temp}
\end{equation}
where $\tau_{init}$ denotes the initial value, $\tau_{range}$ denotes the range for $\tau$, $\delta(\cdot)$ is the sigmoid function, $T_{pred}$ is the predicted value. We default to set $\tau_{init}$ and $\tau_{range}$ to 1 and 20, so that all normal values can be included.

Compared to Global-T, Instance-T can achieve better distillation performance due to its better representation ability. In the following experiments, we mainly use the global version as the default scheme. We demonstrate the effectiveness of the instance-wise temperature method in Table~\ref{table:global_and_instance}.
To get a better understanding of our method, we describe the training procedure in Algorithm~\ref{algo:demo}.

\section{Experiments}

\begin{table}
	\begin{center}
	    \resizebox{0.85\linewidth}{!}{
		\begin{tabular}{cccc}
			\hline\noalign{\smallskip}
			Teacher	    & ResNet-56  & ResNet-110 & WRN-40-2  \\
			Acc		    & 72.34      & 74.31	  & 75.61     \\
			\hline\noalign{\smallskip}
			Student	    & ResNet-20  & ResNet-32  & WRN-40-1  \\
			Acc		    & 69.06      & 71.14	  & 71.98	  \\
			\hline\noalign{\smallskip}
			Vanilla KD  & 70.66      & 73.08      & 73.54     \\
			MACs        & 41.6M      & 70.4M      & 84.7M     \\
			Time        & 10s        & 15s        & 17s       \\
			\hline\noalign{\smallskip}
			Global-T    & 71.19      & 73.52      & 73.93     \\
			MACs        & 41.6M      & 70.4M      & 84.7M     \\
			Time        & 10s        & 15s        & 17s       \\
			\hline\noalign{\smallskip}
			Instance-T  & 71.32      & 73.61      & 74.10     \\
			MACs        & 41.7M      & 70.5M      & 84.8M     \\
			Time        & 11s        & 17s        & 18s       \\
			\hline
		\end{tabular}
		}
		\caption{
			Comparison of global and instance-wise CTKD with various backbones on CIFAR-100. 
			“Time": The time required for one epoch of training.
		}
		\label{table:global_and_instance}
	\end{center}
\end{table}

\begin{table*}
	\begin{center}
	    \resizebox{1\linewidth}{!}{
		\begin{tabular}{cccccccc}
			\hline\noalign{\smallskip}
			Teacher	& ResNet-56            & ResNet-110            &  ResNet-110            & WRN-40-2              & WRN-40-2           
			& ResNet32x4 & ResNet32x4\\
			Acc	    & 72.34	   	            & 74.31	                & 74.31                 & 75.61                 & 75.61     
			& 79.42  & 79.42\\
			\hline\noalign{\smallskip}
			Student	& ResNet-20             & ResNet-32             & ResNet-20             & WRN-16-2              & WRN-40-1 
			& ShuffleNet-V1 & ShuffleNet-V2\\
			Acc     & 69.06	   	            & 71.14	                & 69.06	                & 73.26	                & 71.98    
			& 70.70  &  71.82  \\
			\hline\noalign{\smallskip}
			PKT	& 70.85 $\pm$ 0.22	& 73.36 $\pm$ 0.15		&  70.88 $\pm$ 0.16		& 74.82	$\pm$ 0.19		& 74.01 $\pm$ 0.23
			& 74.39 $\pm$ 0.16      & 75.10 $\pm$ 0.11  \\
			+CTKD	& 71.16 $\pm$ 0.08~\scriptsize{\largeimprove{(+0.31)}}	& 73.53 $\pm$ 0.05~\scriptsize{\minorimprove{(+0.17)}} & 71.15 $\pm$ 0.09~\scriptsize{\largeimprove{(+0.27)}} & 75.32 $\pm$ 0.11~\scriptsize{\largeimprove{(+0.52)}} & 74.11 $\pm$ 0.20~\scriptsize{\minorimprove{(+0.10)}}
			& 74.68 $\pm$ 0.16~\scriptsize{\largeimprove{(+0.29)}}      & 75.47 $\pm$ 0.19~\scriptsize{\largeimprove{(+0.37)}}  \\
			\hline\noalign{\smallskip}
			SP	& 70.84 $\pm$ 0.25   	& 73.09 $\pm$ 0.18	  	& 70.74	$\pm$ 0.23			& 74.88	$\pm$ 0.28		& 73.77 $\pm$ 0.20    & 74.97 $\pm$ 0.28      & 75.59 $\pm$ 0.15 \\
			+CTKD	& 71.27 $\pm$ 0.10~\scriptsize{\largeimprove{(+0.43)}} & 73.39 $\pm$ 0.11~\scriptsize{\largeimprove{(+0.30)}} & 71.13 $\pm$ 0.13 ~\scriptsize{\largeimprove{(+0.39)}}  & 75.33 $\pm$ 0.14~\scriptsize{\largeimprove{(+0.45)}} & 74.00 $\pm$ 0.15~\scriptsize{\largeimprove{(+0.23)}} 
			& 75.37 $\pm$ 0.17~\scriptsize{\largeimprove{(+0.40)}} & 75.82 $\pm$ 0.18~\scriptsize{\largeimprove{(+0.23)}}  \\
			\hline\noalign{\smallskip}
			VID	& 70.62 $\pm$ 0.08   	& 73.02 $\pm$ 0.10	  	& 70.59 $\pm$ 0.19 			& 74.89	$\pm$ 0.16		& 73.60 $\pm$ 0.26    & 74.81 $\pm$ 0.17      & 75.24 $\pm$ 0.05  \\
			+CTKD	& 70.75 $\pm$ 0.11~\scriptsize{\minorimprove{(+0.13)}}	& 73.38 $\pm$ 0.24~\scriptsize{\largeimprove{(+0.36)}}	& 71.09 $\pm$ 0.24~\scriptsize{\largeimprove{(+0.50)}}	& 75.22 $\pm$ 0.20~\scriptsize{\largeimprove{(+0.33)}} & 73.81 $\pm$ 0.24~\scriptsize{\largeimprove{(+0.21)}}
			& 75.19 $\pm$ 0.14~\scriptsize{\largeimprove{(+0.38)}} & 75.52 $\pm$ 0.11~\scriptsize{\largeimprove{(+0.28)}}  \\
			\hline\noalign{\smallskip}
			CRD	&  71.69	$\pm$ 0.15	 	& 73.63	$\pm$ 0.19			&  71.38 $\pm$ 0.04		& 75.53 $\pm$ 0.10		& 74.36 $\pm$ 0.10	& 75.13 $\pm$ 0.33     & 75.90 $\pm$ 0.15 \\
			+CTKD	&  72.11 $\pm$ 0.15~\scriptsize{\largeimprove{(+0.42)}}& 74.10 $\pm$ 0.20~\scriptsize{\largeimprove{(+0.47)}}	& 72.02 $\pm$ 0.10~\scriptsize{\largeimprove{(+0.64)}}		& 75.75 $\pm$ 0.27~\scriptsize{\largeimprove{(+0.22)}} & 74.69 $\pm$ 0.05~\scriptsize{\largeimprove{(+0.33)}}
			& 75.47 $\pm$ 0.22~\scriptsize{\largeimprove{(+0.34)}} & 76.21 $\pm$ 0.19~\scriptsize{\largeimprove{(+0.31)}}  \\
			\hline\noalign{\smallskip}
			SRRL	& 71.13 $\pm$ 0.18		& 73.48 $\pm$ 0.16			& 	71.09 $\pm$ 0.21		& 75.69	$\pm$ 0.19 		& 74.18 $\pm$ 0.03	& 75.36 $\pm$ 0.25     & 75.90 $\pm$ 0.09 \\
			+CTKD	& 71.45 $\pm$ 0.15~\scriptsize{\largeimprove{(+0.32)}} & 73.75 $\pm$ 0.30~\scriptsize{\largeimprove{(+0.27)}}	& 71.48 $\pm$ 0.14~\scriptsize{\largeimprove{(+0.39)}} & 75.96 $\pm$ 0.06~\scriptsize{\largeimprove{(+0.27)}} & 74.40 $\pm$ 0.13~\scriptsize{\largeimprove{(+0.22)}}
			& 75.70 $\pm$ 0.22~\scriptsize{(\largeimprove{+0.34)}} & 76.00 $\pm$ 0.22~\scriptsize{\minorimprove{(+0.10)}}  \\
			\hline\noalign{\smallskip}
			DKD	& 71.43 $\pm$ 0.13		& 73.66 $\pm$ 0.15 		& 71.28 $\pm$ 0.20			& 75.70 $\pm$ 0.06				& 74.54 $\pm$ 0.12	& 75.44 $\pm$ 0.20     & 76.48 $\pm$ 0.08 \\
			+CTKD	& 71.65 $\pm$ 0.24~\scriptsize{\largeimprove{(+0.27)}} & 74.02 $\pm$ 0.29~\scriptsize{\largeimprove{(+0.36)}}	& 71.70 $\pm$ 0.10~\scriptsize{\largeimprove{(+0.42)}} & 75.81 $\pm$ 0.14~\scriptsize{\minorimprove{(+0.11)}}	& 74.59 $\pm$ 0.08~\scriptsize{\minorimprove{(+0.05)}}	
			& 75.93 $\pm$ 0.29~\scriptsize{\largeimprove{(+0.49)}}  & 76.94 $\pm$ 0.04~\scriptsize{\largeimprove{(+0.46)}}  \\
			\hline
		\end{tabular}
    	}
		\caption{
			Top-1 accuracy of the student network on CIFAR-100. 
			\largeimprove{Red} numbers denote non-trivial improvement. \minorimprove{Blue} numbers denote slight improvement.
		}
		\label{table:method_comparison}
	\end{center}
\end{table*}

\begin{table*}[t]
	\begin{center}
	    \resizebox{0.85\linewidth}{!}{
		\begin{tabular}{ccc|cc|cc|cc|cc|cc}
			\hline\noalign{\smallskip}
			          & Teacher & Student & KD    & +CTKD  & PKT    & +CTKD  & RKD    & +CTKD  & SRRL   & +CTKD & DKD    & +CTKD  \\
			\hline\noalign{\smallskip}
			Top-1     & 73.96   & 70.26   & 70.83 & 71.32  & 70.92  & 71.29  & 70.94  & 71.11  & 71.01  & 71.30 & 71.13  & 71.51  \\
            Top-5     & 91.58   & 89.50   & 90.31 & 90.27  & 90.25  & 90.32  & 90.33  & 90.30  & 90.41  & 90.42 & 90.31  & 90.47  \\
            \hline
		\end{tabular}
		}
		\caption{
			Top-1/-5 accuracy on ImageNet-2012. We set ResNet-34 as the teacher and ResNet-18 as the student.
		}
		\label{table:imagenet}
	\end{center}
\end{table*}

We evaluate our CTKD on various popular neural networks e.g., VGG~\cite{simonyan2014very}, ResNet~\cite{he2016deep}~(abbreviated as RN), Wide ResNet~\cite{zagoruyko2016wide}~(WRN), ShuffleNet~\cite{zhang2018shufflenet,ma2018shufflenet}~(SN) and MobileNet~\cite{howard2017mobilenets,sandler2018mobilenetv2}~(MN).
As an easy-to-use plug-in technique, we applied our CTKD to the existing distillation frameworks including vanilla KD~\cite{hinton2015distilling}, PKT~\cite{passalis2018learning}, SP~\cite{tung2019similarity}, VID~\cite{ahn2019variational}, CRD~\cite{tian2019contrastive}, SRRL~\cite{yang2021knowledge} and DKD~\cite{zhao2022decoupled}.
The evaluations are made in comparison to state-of-the-art approaches based on standard experimental settings. All results are reported in means (standard deviations) over 3 trials.

\textbf{Dataset.}
The CIFAR-100 dataset consists of colored natural images with $32\times32$ pixels. The training and testing sets contain 50K and 10K images, respectively. ImageNet-2012~\cite{deng2009imagenet} contains 1.2M images for training, and 50K for validation, from 1K classes. The resolution of input images after pre-processing is $224\times224$. MS-COCO~\cite{lin2014microsoft} is an 80-category general object detection dataset. The train2017 split contains 118k images, and the val2017 split contains 5k images.

\textbf{Implementation details.} All details are attached in supplement due to the page limit.


\begin{figure}[t]
	\centering
	\includegraphics[width=1\linewidth]{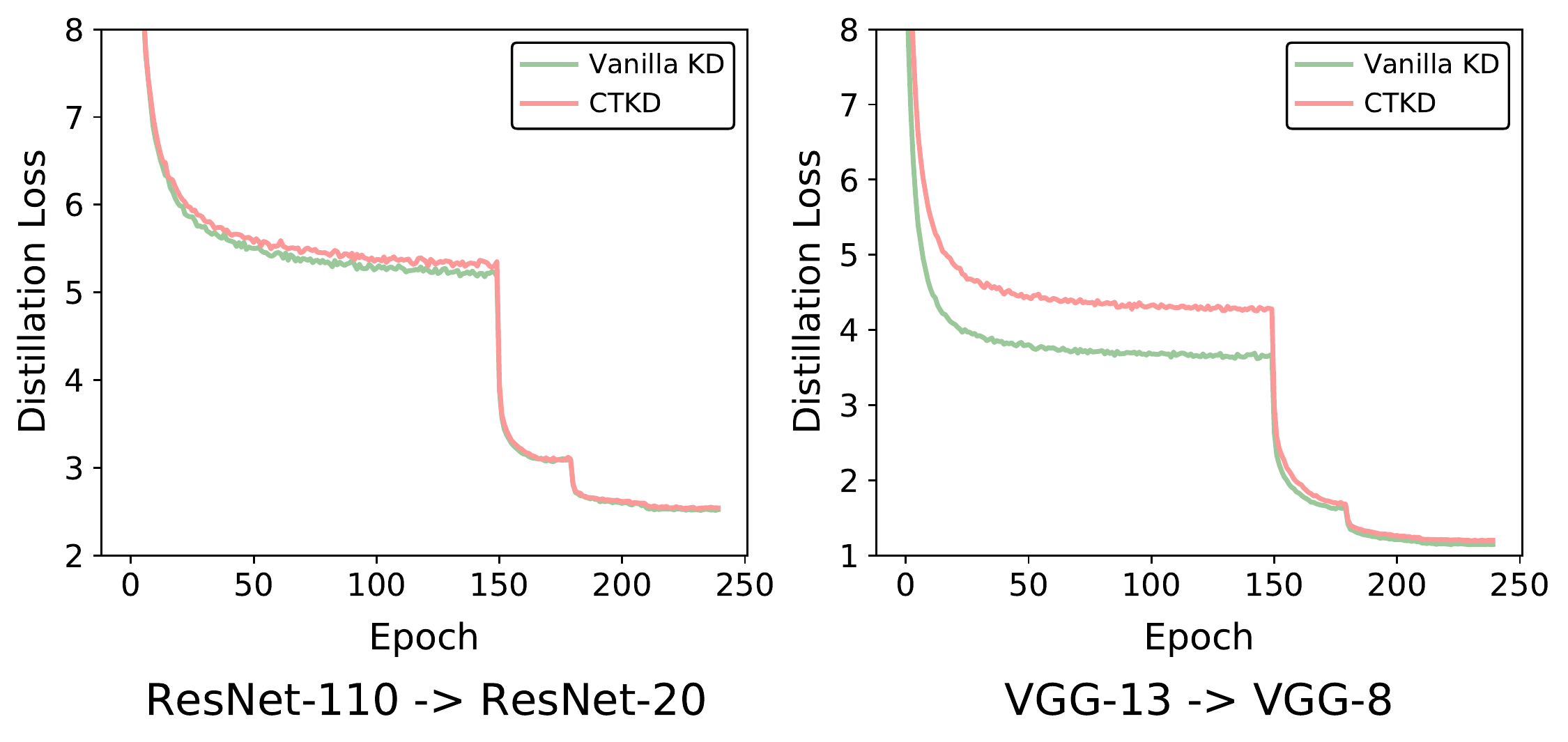}
	\caption{
        The curves of distillation loss during training. Our adversarial distillation technique makes the optimization process harder than the vanilla method as expected. 
	}
	\label{fig:loss_curve}
\end{figure}

\subsection{Main Results} 

\textbf{CIFAR-100 classification.} Table~\ref{table:vanilla_kd_cifar100} shows the top-1 classification accuracy on CIFAR-100 based on eleven different teacher-student pairs. We can observe that all different student networks benefit from our method and the improvement are quite significant in some cases. 

Fig.~\ref{fig:loss_curve} shows the loss curves of vanilla KD and CTKD. During training, the temperature module is optimized to maximize the distillation loss, which satisfies the condition in Eqn.~\eqref{equation:curriculum_growing}.
While the student is optimized to minimize the distillation loss, which plays a leading role in this mini-max game. So the overall losses still show a downward trend. As shown in Fig.~\ref{fig:loss_curve}, the distillation loss of CTKD is higher than the vanilla method, proving the effectiveness of adversarial temperature distillation.
We can observe that the distillation loss of CTKD is higher than that of vanilla KD, proving the effect of the adversarial operation.

\begin{figure}[t]
	\centering
	\includegraphics[width=0.9\linewidth]{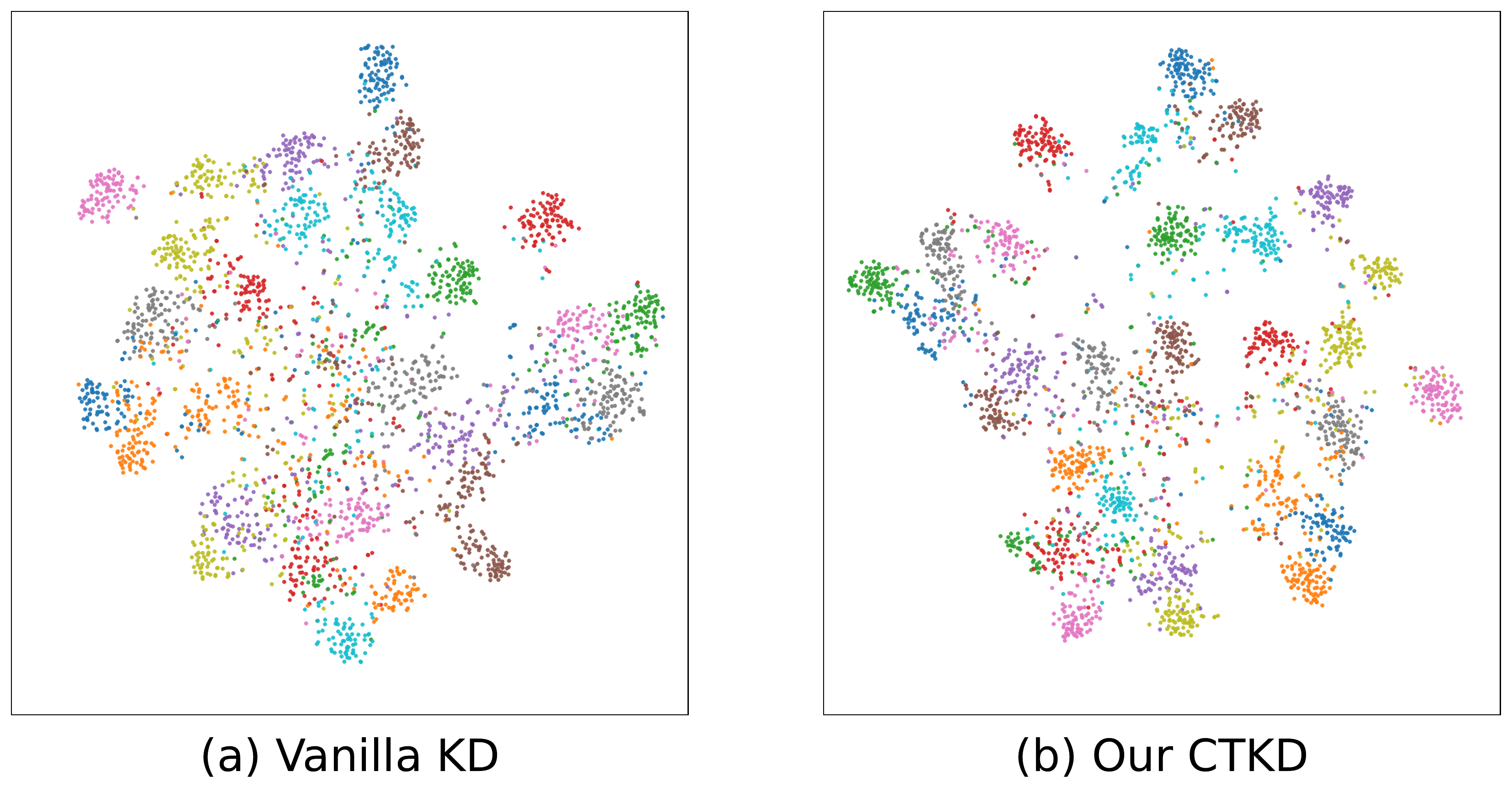}
	\caption{
        t-SNE of features learned by KD and CTKD.
	}
	\label{fig:tsne}
\end{figure}

\begin{figure*}[t]
	\centering
	\includegraphics[width=1\linewidth]{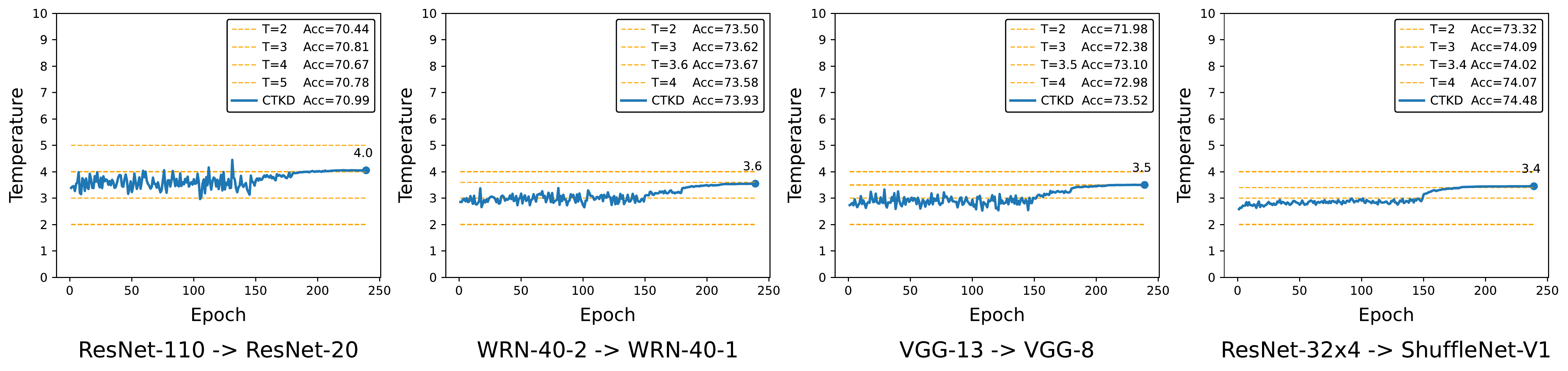}
	\caption{
        The learning curves of temperature during training. The yellow dotted line represents the vanilla distillation method at a specified fixed temperature. The solid blue line represents the dynamic temperature learning process. Our dynamic curriculum temperature outperforms the static method.
    }
	\label{fig:temp_curve}
\end{figure*}

Fig.~\ref{fig:tsne} demonstrates that representations of our method are more separable than vanilla KD, proving that CTKD benefits the discriminability of deep features. Fig.~\ref{fig:temp_curve} shows the learning curves of temperature during training.
Compared to fixed temperature distillation, our curriculum temperature method achieves better results via an effective dynamic mechanism.

\textbf{Global and instance-wise temperature.}
Table~\ref{table:global_and_instance} shows the top-1 classification accuracy and computational efficiency~(MACs, Time) of the global and instance-wise versions. 
Since the instance-wise method introduces an additional network~(i.e., 2-layer MLP) to obtain stronger representation ability, it requires more computational cost than the global version. 
From this table, we can see that both versions can improve student performance at a negligible additional computational cost. 
We mainly use the global version in the following experiments.

\textbf{Applied to existing distillation works.}
As an easy-to-use plug-in technique, CTKD can be seamlessly integrated into existing distillation works. 
As shown in Table~\ref{table:method_comparison}, our method brings comprehensive improvements to six state-of-the-art methods based on seven teacher-student pairs. 
More importantly, CTKD does not incur additional computational costs to the methods since it only contains a lightweight learnable temperature module and a non-parameterized GRL.

\textbf{ImageNet-2012 classification.}
Table~\ref{table:imagenet} reports the top-1/-5 accuracy of image classification on ImageNet-2012. As a plug-in technique, we also applied our CTKD to four existing state-of-the-art distillation works. 
The result shows that CTKD can still works on the large-scale dataset effectively.

\begin{table}[h]
	\begin{center}
	    \resizebox{0.99\linewidth}{!}{
		\begin{tabular}{ccccccc}
			\hline\noalign{\smallskip}
			     &  mAP  & AP50  & AP75  & APl   & APm   & APs   \\
			\hline\noalign{\smallskip}
			T: RN-101 & 42.04 & 62.48 & 45.88 & 54.60 & 45.55 & 25.22 \\
			S: RN-18  & 33.26 & 53.61 & 35.26 & 43.16 & 35.68 & 18.96 \\
			\hline\noalign{\smallskip}
			KD   & 33.97 & 54.66 & 36.62 & 44.14 & 36.67 & 18.71 \\
			+CTKD & 34.56 & 55.43 & 36.91 & 45.07 & 37.21 & 19.08 \\
			\hline\hline\noalign{\smallskip}
			T: RN-50  & 40.22 & 61.02 & 43.81 & 51.98 & 43.53 & 24.16 \\
            S: MN-V2  & 29.47 & 48.87 & 30.90 & 38.86 & 30.77 & 16.33 \\
            \hline\noalign{\smallskip}
            KD   & 30.13 & 50.28 & 31.35 & 39.56 & 31.91 & 16.69 \\
            +CTKD & 31.39 & 52.34 & 33.10 & 41.06 & 33.56 & 18.15 \\
            \hline
		\end{tabular}
		}
		\caption{
			Results on MS-COCO based on Faster-RCNN~\cite{ren2015faster}-FPN~\cite{lin2017feature}: AP evaluated on val2017. 
		}
		\label{table:coco_results}
	\end{center}
\end{table}

\textbf{MS-COCO object detection.}
We also apply our method to the object detection task. We follow the object detection implementation of DKD. As shown in Table~\ref{table:coco_results}, our CTKD can further boost the detection performance.

\subsection{Ablation Study}
 In the following experiments, we evaluate the effectiveness of hyper-parameters and components on CIFAR-100. We set ResNet-110 as the teacher and ResNet-32 as the student.

\textbf{Curriculum parameters.} 
Table~\ref{table:curriculum_params} reports the student accuracy with different $\lambda_{min}, \lambda_{max}$, and $E_{loops}$. Table~\ref{table:fixed_lambda} reports the distillation results with different fixed $\lambda$. 
The training of students needs to gradually increase the learning difficulty. Directly starting with a fixed high-difficulty task will significantly reduce the performance of students, especially when $\lambda$ is greater than 4. Besides, as shown in the sixth and seventh columns of Table~\ref{table:curriculum_params}, rapidly increasing parameter $\lambda$ in a short time can also be detrimental to student training. When we smooth the learning difficulty of the student and increase $E_{loops}$, the performance can be further improved.

\begin{table}
	\begin{center}
	    \resizebox{0.9\linewidth}{!}{
		\begin{tabular}{cccccccc}
			\hline\noalign{\smallskip}
			\multirow{2}*{$E_{loops}$} & \multicolumn{5}{c}{[$\lambda_{min}$, ${\lambda_{max}}$]} \\
			\noalign{\smallskip}
			~                          & [0, 1]         & [0, 2] & [0, 5]            & [0, 10]  & [1, 10] \\
			\hline\noalign{\smallskip}
			10 Epoch                   & \cellcolor{lightgray!30}\textbf{73.52} & 73.16  & 73.12 	         & 73.05    & 72.58  \\
			20 Epoch                   & \cellcolor{lightgray!30}73.44	        & \cellcolor{lightgray!30}73.48  & 73.01	         & 73.00    & 72.88  \\
			40 Epoch                   & 73.26	        & \cellcolor{lightgray!30}73.40  & \cellcolor{lightgray!30}\underline{73.50} & 73.15    & 72.95  \\
			80 Epoch                   & 73.35  	    & \cellcolor{lightgray!30}73.46  & \cellcolor{lightgray!30}\textbf{73.52}    & \cellcolor{lightgray!30}73.41    & 73.12  \\
			120 Epoch                  & 73.31          & 73.39  & 73.16             & \cellcolor{lightgray!30}73.36    & 73.04  \\
			240 Epoch                  & 73.23          & 73.29  & 73.20             & \cellcolor{lightgray!30}73.42    & 73.08  \\
			\hline
		\end{tabular}
		}
		\caption{
			Range of dynamic curriculum $\lambda$. Smoothly increasing task difficulty is beneficial to students' learning.
		}
		\label{table:curriculum_params}
	\end{center}
\end{table}

\begin{table}
	\begin{center}
	    \resizebox{1\linewidth}{!}{
		\begin{tabular}{cccccc|c}
			\hline\noalign{\smallskip}
			Fixed $\lambda$  &  1    &  2    & 4     &    5   &   10  & Curriculum \\
			\hline\noalign{\smallskip}
			Acc              & 73.26 & 73.36 & 73.16 & 72.78 & 72.82 & \textbf{73.52}   \\
			\hline
		\end{tabular}
		}
		\caption{
			Training with fixed $\lambda$. Compared with curriculum $\lambda$, 
			directly training the student with the fixed high-difficulty task (e.g., $\lambda \textgreater 4$) will reduce distillation performance.
		}
		\label{table:fixed_lambda}
	\end{center}
\end{table}


\textbf{Curriculum strategy.} 
In Table~\ref{table:curriculum_strategy}, we compare the performance of different curriculum strategies. 
``None$_{\tau=1}$, 10 Epoch" means that during the first 10 epochs of training, we only use vanilla distillation, and set the temperature $\tau=1$. After 10 epochs, we start to train the student with CTKD, and $\lambda$ is fixed to 1.
``Lin$_{[0,1]}$, 10 Epoch" means that we use CTKD to train the student with a linear increasing strategy. The parameter $\lambda$ is gradually increased from 0 to 1 in 10 epochs of training. The value of $\lambda$ remains 1 until the end. From this table, we can see that the cosine curriculum strategy works the best.

\begin{table}[t]
	\begin{center}
	    \resizebox{0.89\linewidth}{!}{
		\begin{tabular}{ccccc}
			\hline\noalign{\smallskip}
			\multirow{2}*{$E_{loops}$} & \multicolumn{4}{c}{Curriculum Strategy} \\
			~        & None$_{\tau=1}$ & None$_{\tau=4}$  & Lin$_{[0,1]}$ & Cos$_{[0,1]}$  \\
			\noalign{\smallskip}\hline\noalign{\smallskip}
			10 Epoch & 73.21	       & 73.07 	          & 73.31         & \textbf{73.52} \\
			20 Epoch & 73.24	       & 73.06	          & 73.45         & 73.44          \\
			40 Epoch & 73.33	       & 73.07	          & 73.10         & 73.26          \\
			\hline
		\end{tabular}
		}
		\caption{
			Comparison of different curriculum strategies. Cosine curriculum strategy works the best.
		}
		\label{table:curriculum_strategy}
	\end{center}
\end{table}

\textbf{Adversarial temperature and curriculum distillation.} 
We evaluate the effectiveness of these two elements as shown in Table~\ref{table:components}.
The second row means that we only adopt the adversarial temperature technique and use the fixed learning difficulty~(i.e., fix $\lambda$ to 1) to train the student. The results demonstrate that learning the temperature parameter in an adversarial manner can also improve distillation performance. The third row shows that the cooperation of two elements can achieve better results than a single element.

\begin{table}[t]
	\begin{center}
	    \resizebox{0.99\linewidth}{!}{
		\begin{tabular}{cccccc}
			\hline\noalign{\smallskip}
			\multirow{2}*{AT}   & \multirow{2}*{CD}  & ResNet-56  & ResNet-110 & WRN-40-2 & VGG-13 \\
			 ~                  & ~                  & ResNet-20  & ResNet-32  & WRN-16-2 & VGG-8  \\
			\hline\noalign{\smallskip}
			                     &                   & 70.66      & 73.08      & 74.92    & 72.98  \\
			\checkmark           &                   & 71.01      & 73.26      & 74.99    & 73.43  \\
			\checkmark           & \checkmark        & 71.19      & 73.52      & 75.45    & 73.52  \\
			\hline
		\end{tabular}
		}
		\caption{
			Ablation of Adversarial Temperature~(AT) module and Curriculum Distillation~(CD) strategy. The first row indicates the vanilla distillation performance.
		}
		\label{table:components}
	\end{center}
\end{table}

\section{Conclusion}

In this paper, we propose a curriculum-based distillation approach, termed Curriculum Temperature for Knowledge Distillation, which organizes the distillation task from easy to hard through a dynamic and learnable temperature.
The temperature is learned during the student’s training process with a reversed gradient that aims to maximize the distillation loss (i.e., increase  the learning difficulty) between teacher and student in an adversarial manner. 
As an easy-to-use plug-in technique, CTKD can be seamlessly integrated into existing state-of-the-art knowledge distillation frameworks and brings general improvements at a negligible additional computation cost. 

\section{Acknowledgments}
This work was supported by the Young Scientists Fund of the National Natural Science Foundation of China~(Grant No.62206134).

\bibliography{aaai23.bib}

\clearpage

\setcounter{section}{0}
\section{Supplementary Materials}

\subsection{Implementation Details}

All the methods are implemented by PyTorch~\cite{paszke2019pytorch}. 
\textbf{CIFAR-100:} We follow the standard data augmentation scheme for all training images as in~\cite{chen2021distilling,zhao2022decoupled}, i.e. random cropping and horizontal flipping.
We use the stochastic gradient descents~(SGD) as the optimizer with momentum 0.9 and weight decay 5e-4. The learning rate is divided by 10 at 150, 180, and 210 epochs, for a total of 240 epochs. The mini-batch size is 64.
The initial learning rate is set to 0.01 for MobileNet/ShuffleNet architectures and 0.05 for other architectures. We set $\lambda_{max}$, $\lambda_{min}$ and $E_{loops}$ to $1$, $0$ and $10$. We use the Global-T version as the default scheme in all experiments unless we specified. 
The temperature T in existing KD methods is default set to 4 in all experiments. 
In Table~\ref{table:vanilla_kd_cifar100}, we directly cite the vanilla KD results reported in~\cite{chen2021distilling,zhao2022decoupled} since we use the same experimental settings. 
In Table~\ref{table:method_comparison}, we combine existing methods~(i.e., PKT, SP, VID, CRD, SRRL) with KD and rerun the experiments according to their official implementations~\footnote{https://github.com/HobbitLong/RepDistiller}\footnote{https://github.com/jingyang2017/KD\_SRRL}\footnote{https://github.com/megvii-research/mdistiller} for fair comparison.

\textbf{ImageNet:} We set the initial learning rate to 0.1 and divide the learning rate by 10 at 30, 60, and 90 epochs. We follow the standard training process but train for 20 more epochs~(i.e., 120 epochs in total). Weight decay is set to 1e-4. $E_{loops}$ is set to 5. The mini-batch size is 256. In Table~\ref{table:imagenet}, all teacher and student models are retrained based on 120 epochs.

\textbf{MS-COCO:} Our implementation for MS-COCO follows the settings in~\cite{chen2021distilling,zhao2022decoupled}. We adopt the two-stage object detection method Faster-RCNN~\cite{ren2015faster} with FPN~\cite{lin2017feature} as the basic network. In our CTKD implementation, we set $\lambda_{max}$, $\lambda_{min}$ and $E_{loops}$ to $0.25$, $0$ and $6$, respectively.

\textbf{Training details:}
Existing methods have designed various forms of knowledge transfer in addition to the traditional soft labels. It also contains representation-based~\cite{yim2017gift,chen2021distilling} and relationship-based~\cite{park2019relational,peng2019correlation} methods. For a modern distillation task, the training objective for different methods can be simply formulated as follows:
\begin{equation}
\begin{aligned}
    L_{total} = \alpha L_{cross-entropy} + & ~ \beta L_{kl-divergence} \\ + & ~ \gamma L_{distill-forms}.
\end{aligned}
\end{equation}
where $\alpha$, $\beta$ and $\gamma$ are balancing hyper-parameters.

For CIFAR-100, we default set $\alpha=0.1$ and $\beta=0.9$. For ImageNet, we default set $\alpha=1$ and $\beta=1$.
The results in Table~\ref{table:method_comparison} and Table~\ref{table:imagenet} are reproduced according to the official implementation with the following hyper-parameters:
\begin{itemize}
    \item PKT~\cite{passalis2018learning}: $\gamma=30000$.
    \item SP~\cite{tung2019similarity}: $\gamma=3000$.
    \item VID~\cite{ahn2019variational}: $\gamma=1$.
    \item CRD~\cite{tian2019contrastive}: $\gamma=0.8$.
    \item SRRL~\cite{yang2021knowledge}: $\gamma=1$.
    \item DKD~\cite{zhao2022decoupled}: $\gamma=1$. For CIFAR-100 and ImageNet, we set $\alpha=1$ and $\beta=0$.
\end{itemize}

\subsection{Additional Experiments}

\textbf{Comparison with grid search results.}
Since our CTKD is a dynamic temperature learning method, we further compare the performance with the fixed temperature distillation method. Table~\ref{table:grid_search} details the performance under different fixed temperature values. We can observe that the optimal temperature value varies widely for different teacher-student pairs. Choosing the default temperature value~(i.e., 4) is not the best option for existing works, and performing a grid search for the optimal parameter will significantly increase the training complexity. By using our CTKD, the temperature value can be automatically learned and dynamically changed according to different learning stages. The bold results demonstrate the effectiveness of our method.

\begin{figure}[t]
	\centering 
	\includegraphics[width=1\linewidth]{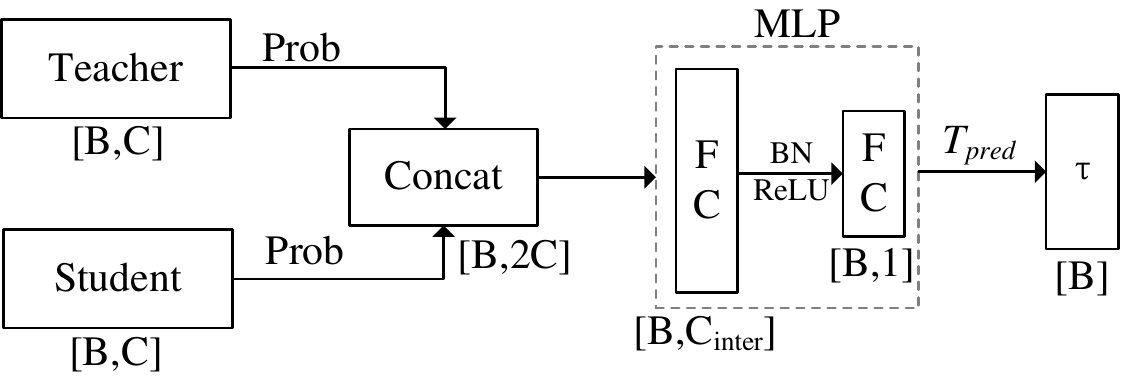}
	\caption{
		The detailed architecture of our instance-wise temperature method. $B$ denotes the batch size, $C$ denotes the number of classes, and $C_{inter}$ denotes the number of intermediate channels in the MLP.
	}
	\label{fig:instance}
\end{figure}

\begin{table}[t]
	\begin{center}
	    \resizebox{0.85\linewidth}{!}{
		\begin{tabular}{cccc}
			\hline\noalign{\smallskip}
			Teacher	& ResNet-110 & WRN-40-2 & VGG-13    \\
			Acc		& 74.31	   	 & 75.61	& 74.64     \\
			\hline\noalign{\smallskip}
			Student	& ResNet-20  & WRN-40-1 & VGG-8     \\
			Acc		& 69.06	   	 & 71.98	& 70.36	    \\
			\hline\noalign{\smallskip}
			T=1		& 69.68 	 & 71.90    & 70.42   	\\
			T=2	    & 70.44      & 73.50 	& 71.98	    \\
			T=3	    & \underline{70.81}      & 73.62	& 72.38	    \\
			T=4	    & 70.67      & 73.58    & 72.98  	\\
			T=5	    & 70.78      & 73.66 	& \underline{73.42} 	\\
			T=6	    & 70.50      & 73.73 	& 73.37 	\\
			T=8	    & 70.35      & \underline{73.75}  	& 73.31 	\\
			\hline\noalign{\smallskip}
			CTKD    & \textbf{70.99}& \textbf{73.93} & \textbf{73.52} \\
			\hline
		\end{tabular}
		}
		\caption{
			Comparison to grid search results. Bold and underline denote the best and second best results. 
		}
		\label{table:grid_search}
	\end{center}
\end{table}

\begin{table}[t]
	\begin{center}
	    \resizebox{0.97\linewidth}{!}{
		\begin{tabular}{ccccccc}
			\hline\noalign{\smallskip}
			$C_{inter}$  &  64    &  128   & 256     &   512   &   1024  & 2048 \\
			\hline\noalign{\smallskip}
			Acc          & 71.28  & 71.31  & \textbf{71.32}   & 71.26   & 71.21   & 71.28 \\
			\hline
		\end{tabular}
		}
		\caption{
			Comparison of different intermediate channels in instance-wise method. We set ResNet-56 as the teacher and ResNet-20 as the student.
		}
		\label{table:channels}
	\end{center}
\end{table}

\textbf{Intermediate channels of instance-wise temperature module.}
Fig.~\ref{fig:instance} shows the detailed architecture of our instance-wise CTKD. A 2-layer MLP is introduced in our method.
In Table~\ref{table:channels}, we compare the performance of different intermediate channels on CIFAR-100. 
Since our instance-wise method needs to predict an appropriate temperature value for each instance, it requires stronger representational capabilities than the global method. From this table, we can see that a larger intermediate channel~(e.g., $C_{inter}\geq 256$) works better for the instance-wise method. 

\subsection{Discussion}

\textbf{Detailed comparison with MKD.}
The previous work MKD~\cite{liu2022meta} shares the same goal as our proposed CTKD, which is to learn an appropriate temperature parameter for the distillation process. But there are big differences between these two methods:
\begin{itemize}
    \item The learning objectives and optimization directions of these two methods are different. In MKD, the temperature module is trained to \textbf{minimize} the L2-norm loss between the \textbf{student}'s outputs and \textbf{ground truth} labels on the extra \textbf{validation set}. But CTKD proposes to adversarially learn the temperature value that \textbf{maximizes} the KL-divergence loss between the \textbf{teacher}'s outputs and the \textbf{student}'s outputs on the \textbf{training set}. 
    \item The datasets used for training are different. MKD needs to split and preserve an extra validation set to train the temperature module. CTKD is more efficient than MKD since our method can be trained directly on the training set without any additional operations.
    \item The working conditions are different. MKD mainly focuses on the strong data augmentation condition while most existing KD methods work under normal augmentation. Directly combining MKD with existing distillation methods under strong augmentation may cause severe performance degradation~\cite{das2020empirical}. In contrast, our CTKD works under normal circumstances. It can be seamlessly integrated into most existing works and achieve comprehensive improvements.
\end{itemize}

Note that MKD hasn’t released its official code and we don't know the detailed training settings. We have tried to re-implement MKD but we failed to achieve the accuracy reported in the paper. So now we cannot provide accurate experimental comparisons. After MKD releases the official code, we will update the experimental result in time.

\textbf{Comparison with existing teacher-student curriculum learning framework.}
In curriculum learning, teachers are introduced to determine the optimal curriculum strategy for students, e.g., assigning different weights to the training samples~\cite{jiang2018mentornet, kim2018screenernet} or arranging the learning sequence of sub-tasks~\cite{florensa2017reverse}. 
In knowledge distillation, teachers are mainly used as the learning target. The student is trained to minimize the representation (e.g. soft labels~\cite{hinton2015distilling} or features~\cite{romero2014fitnets}) distance from the teacher. After the distillation, the student can master the expertise of the teacher.
The teacher model in both methods shares the same goal, which aims to assist the target (i.e., student) model to achieve better performance. But these two frameworks have different learning targets and training paradigms, as we described above.

\textbf{Differences between curriculum and warmup strategy.}
The warmup method gradually increases the hyper-parameter~(e.g., learning rate) over a few epochs until it reaches the constant final value. It only works at the beginning of training. However, our curriculum strategy can work throughout the training process. As shown in paper Table~\ref{table:curriculum_params}, our strategy still works well when we use a larger difficulty range and longer training epochs~(i.e., 240 of 240 epochs in total).

\end{document}